# Hyacinth6B: A Large Language Model for Traditional Chinese

**Chih-Wei Song**  **Yin-Te Tsai**

Computer Science and Communicate Engineering

Providence University

**Abstract**

Since the emergence of ChatGPT at the end of 2022, the field of artificial intelligence has been entering a new era. This development not only signifies significant advancements in natural language processing technology but also profoundly impacts the domain of generative artificial intelligence, including image, video, and speech generation. The flourishing development of Large Language Models (LLMs) has been a key driving force in this transformation. However, it is noteworthy that the domain of Traditional Chinese language models has less research, causing this linguistic community to lag in this wave of innovation. This primary motivation of this study is to address the high hardware and computational demands typically associated with LLMs. Therefore, our goal is to find a balance between model lightness and performance, striving to maximize performance while using a comparatively lightweight model. Hyacinth6B was developed with this objective in mind, aiming to fully leverage the core capabilities of LLMs without incurring substantial resource costs, effectively pushing the boundaries of smaller models' performance. The training approach involves parameter-efficient fine-tuning using the Low-Rank Adaptation (LoRA) method.

At last, we will evaluate Hyacinth6B, examining its performance across various aspects. Hyacinth6B shows commendable performance in certain metrics, even surpassing ChatGPT in two categories. We look forward to providing more resources and possibilities for the field of Traditional Chinese language processing. This research aims to expand the research scope of Traditional Chinese language models and enhance their applicability in different scenarios.

1. Introduction

The emergence of ChatGPT[1] signifies a pivotal moment in the artificial intelligence landscape. This advancement in natural language processing technology has significantly improved the functionality of conversational systems. Focusing more on language model development has spurred significant progress in understanding and generating human language, pushing the boundaries of machine interaction. This breakthrough primarily involves the enhancement of text-based applications, demonstrating how AI can more effectively comprehend, interpret, and respond to human language. This shift has had a profound impact on various sectors, ranging from customer service automation to sophisticated data analysis, emphasizing the transformative power of advanced language models in AI.

Due to the success of ChatGPT, the training strategy of fine-tuning with instructions[2] has also been endorsed by researchers in the language model domain. This study adheres to this strategy in its implementation. Initially, a high-quality pre-trained model, also known as the foundational model, is essential. After a thorough selection process, we have chosen ChatGLM3-6B-base, which was developed by Beijing Tsinghua University KEG lab and Zhipu AI, as the foundational model.

In the process of instruction fine-tuning, we opted for the Low-Rank Adaptation (LoRA) method, a decision influenced by hardware constraints and the method's efficiency. LoRA is a popular training strategy in the LLM field, not only demands minimal computational resources but also achieves performance competitive with full model fine-tuning. This balance between hardware feasibility and effectiveness makes LoRA an ideal choice for our implementation needs. During LoRA training, only the dimensionality reduction matrix and the dimensionality increase matrix are trained, while the parameters of the original model are frozen. For fine-tuning, we choose Traditional Chinese instructions, which was used to train Taiwan LLM, developed by Miulab in National Taiwan University. This model utilizes a Traditional Chinese dataset during the instruction tuning process, providing us with a rich foundation of data.

Through this research, our aim is to provide more advanced tools and resources to propel the application and development of Traditional Chinese language processing. The purpose of this study is to expand the research scope of Traditional Chinese language models, enhance their application value in various scenarios, and contribute to the future of this field.

2. Related Work

 2.1 ChatGLM3

ChatGLM3[3, 4] is a new generation of bilingual pre-trained conversational models for Chinese and English, launched in 2023 by Zhipu AI and Beijing Tsinghua University KEG Lab. The open-source version, ChatGLM3-6B, retains many of excellent characteristics of its predecessors, such as smooth conversation flow and low deployment threshold. ChatGLM3 introduces the following features:

**More Powerful Base Model**: ChatGLM3-6B-base utilizes a more diverse training dataset, more comprehensive training steps, and a more rational training strategy. In terms of semantics, mathematics, reasoning, code, and knowledge, this model demonstrates outstanding performance. As of October 27, 2023, ChatGLM3-6B-Base is recognized as the strongest performer among foundational models under 10B parameters.

**More Comprehensive Functional Support**: ChatGLM3-6B adopts a newly designed Prompt format, which enhances the capabilities of ChatGLM3-6B comprehensively without impacting the model's general abilities. Additionally, it natively supports complex scenarios such as Function Call, Code Interpreter, and Agent tasks, beyond normal multi-turn dialogues.

**More Extensive Open-Source Sequence**: Besides the conversational model ChatGLM3-6B, the basic model ChatGLM3-6B-Base and the long-text conversation model ChatGLM3-6B-32K are also open-sourced. All these weights are fully accessible for academic research, after registration via a questionnaire and are also permitted for free commercial use.

ChatGLM3-6B offers two different deployment methods. The first is a chat-only conversation mode, while the second integrates the conversation mode with tool models and a code interpreter model. In this study, we only used the first chat conversation mode for deployment and training.

**2.2 Low-Rank Adaptation**

In the field of Parameter-Efficient Fine-Tuning (PEFT) methods, Low-Rank Adaptation (LoRA)[5] stands out as a noteworthy approach, especially in the context of Large Language Models (LLMs). LoRA operates by inserting small, trainable low-rank matrices into each layer of a pre-trained model, such as GPT series or LLaMA[6], effectively fine-tuning the model while keeping the majority of the pre-trained parameters frozen. This technique significantly reduces the number of parameters that need to be trained, thereby decreasing computational costs and memory requirements, yet still manages to achieve performance comparable to full-parameter fine-tuning.

The primary advantage of LoRA lies in its ability to efficiently adapt large-scale models for specific tasks under computational constraints. Unlike conventional fine-tuning methods, which require adjustments to all model parameters, LoRA targets only a small subset, making it an ideal choice for applications with limited computational resources. This efficiency makes LoRA particularly suitable for fine-tuning LLMs in task-specific scenarios, where the deployment environment may not allow for extensive computational overhead.

In comparison to other PEFT methods like P-tuning[7], which focuses on tuning prompt-based parameters, LoRA distinguishes itself by directly modifying the internal weights of the model in a low-rank manner. While P-tuning is effective for leveraging prompts to guide model behavior, LoRA offers a more direct approach to adjusting model behavior by altering its internal representations. This difference in approach allows LoRA to be more versatile in a wider range of tasks and settings, potentially offering better integration and fine-tuning effectiveness for LLMs.

The applications of LoRA in the LLM domain are diverse, ranging from natural language understanding and generation to more complex reasoning and computation tasks. By enabling efficient and effective fine-tuning, LoRA represents a significant advancement in the field of language models, offering a practical solution to harness the power of LLMs in a resource-constrained environment.

**2.3 Traditional Chinese Large Language Models**

The development of language models for Chinese, particularly Traditional Chinese, has historically lagged behind their English counterparts. This is partly due to the lack of robust datasets and less investment in Large Language Model technologies for Chinese languages. Furthermore, the components necessary for Chinese language processing, such as tokenizers and embeddings, differ significantly from those used in English, presenting unique challenges. These differences have resulted in a relative scarcity of Chinese resources in the open-source community.

Despite these challenges, there are ongoing efforts to develop Traditional Chinese language models. One such project is the Taiwan LLM[8], developed by scholars Yen-Ting Lin and Yun-Nung Chen from the National Taiwan University Miu lab. This model is based on the LLaMA architecture and is trained on a vast corpus of Traditional Chinese texts. In our study, we also utilized their open-source dataset. Another notable project is from MediaTek Research, which has released open-source versions of its models BLOOM with 1B1 and 3B parameters[9]. Their latest Traditional Chinese model is Model 7 C. Additionally, MediaTek Research provides the TC-eval evaluation suite[10], containing multiple domain-specific tasks to assess the capabilities of LLMs.

While the development of Traditional Chinese language models faces distinct challenges, including the need for specialized components and a lack of comprehensive datasets, significant strides are being made. Projects like Taiwan LLM and MediaTek's Model 7 C demonstrate the growing interest and effort in enhancing Traditional Chinese language processing capabilities. However, the field still faces potential challenges in terms of resource availability and technological advancement to match the efficacy seen in English language models.

3. **Methodology**

### 3.1 Choice of Foundation Model: ChatGLM3-base

The selection of ChatGLM3-base as the foundational model for this study was based on its outstanding performance and efficiency. Recognized for its superior Chinese language processing under 10 billion parameters, it boasts rapid token generation, averaging 17-18 ms per token on a single RTX 4090. Despite its primary focus on Simplified Chinese, our study leveraged Supervised Fine-Tuning with a Traditional Chinese corpus to potentially eliminate Simplified Chinese outputs. Crucially, ChatGLM3-base is a lightweight model, requiring only about 14G of VRAM for inference without quantization. This makes it an ideal choice for resource-constrained environments, offering small businesses and institutions an efficient solution for training private models without sacrificing precision.

### 3.2 Fine-Tuning Method: LoRA

While the ChatGLM series predominantly employs P-Tuning for fine-tuning, our study opts for Low-Rank Adaptation (LoRA) as our fine-tuning method of choice. The rationale behind this selection is twofold. Firstly, LoRA's lower hardware requirement and its efficiency in fine-tuning make it a more viable option, particularly in resource-constrained environments. Secondly, LoRA offers distinct advantages over P-Tuning in terms of model adaptability and parameter efficiency. Unlike P-Tuning, which focuses on prompt-based parameter adjustments, LoRA modifies the internal weight matrices of the model in a low-rank manner. This allows for more direct and effective modification of the model's behavior, making it particularly suited for tasks requiring nuanced understanding and response generation. LoRA's ability to fine-tune models with a relatively smaller set of trainable parameters without significant loss in performance is a key factor in its selection for this study.

### 3.3 Fine-Tuning Dataset: Instruct Tuning

The fine-tuning process employs the Instruct Tuning approach, structured as single or multi-turn dialogues, each comprising a user query and an assistant response. This method is designed to enable the language model to learn domain-specific knowledge and appropriate response styles to various queries, aligning with human expectations. Instruct Tuning is one of the hottest trends in the LLM domain, inspired by the monumental success achieved by GPT using this methodology.

### 3.4 Dataset Source

The dataset used is the "Traditional Chinese instructions" provided open-source by the National Taiwan University Miu lab. It is a compilation of various datasets such as ShareGPT, Baize, Flan V2, CoT, GPT4-Alpaca, etc., translated into Traditional Chinese. The dataset encompasses categories like contextual Q&A, world knowledge, summarization of long texts, classification, and table comprehension. Prior to use, we performed data cleansing within our capabilities, including deduplication, removal of missing QA pairs, and incorrect data, ultimately cleansing about 10% of the content, amounting to approximately 486,000 entries.

### 3.5 Training Process

The training was conducted on a single RTX 4090, employing LoRA for fine-tuning. The hyperparameter settings were shown in Table 1.

Table1: hyperparameter configuration

| Parameter | Setting |
| --- | --- |
| Batch Size | 8 |
| Gradient Accumulation Steps | 4 |
| Learning Rate | 5e-5 |
| Epochs | 3 |
| LoRA r | 16 |

Training required approximately 20.6GB of VRAM without any quantization (default fp16) and a total of 369 hours in duration. An in-depth analysis and discussion of the model's capabilities and results will be presented in the following chapter.

## 4. Experiment Results

After the implementation of Hyacinth6B, we selected several benchmarks to test the performance of the model and conducted individual capability comparisons with other models to understand the diverse domain abilities of Hyacinth6B. Firstly, we utilized common assessment suites like MMLU[11], as well as CMMLU[12] and C-eval[13], which are primarily focused on Chinese, to test our model's performance. For detailed results, please refer to Tables 2 to 4.

Table2 : MMLU 5-shot Results

| Models | STEM | SocialScience | Humanities | Others | Average |
|---|---|---|---|---|---|
| GPT4 | N/A | N/A | N/A | N/A | 86.50 |
| GPT-3 | 41.40 | 63.90 | 52.50 | 57.90 | 53.90 |
| LLaMA 13B | 35.80 | 53.80 | 45.00 | 53.30 | 46.90 |
| LLaMA 7B | 30.50 | 38.30 | 34.00 | 38.10 | 35.10 |
| Hyacinth6B | **45.90** | **68.61** | 52.37 | **62.73** | **56.93** |

From Table 2, it is evident that Hyacinth6B performs exceptionally well across all areas, particularly in the field of social sciences. Since GPT-4 only provides an overall score in its paper, the detailed scores for individual subjects are unknown[14]. Hyacinth6B trails only behind GPT-4 in terms of average total score and even surpasses LLaMA 13B, which has a larger number of parameters.

Table3 : CMMLU 5-shot Results

| Models | STEM | SocialScience | Humanities | Others | Average |
|---|---|---|---|---|---|
| GPT4 | 65.23 | 72.06 | 72.11 | 74.79 | 70.95 |
| ChatGPT | 47.81 | 56.50 | 55.68 | 62.66 | 55.51 |
| Baichuan-13B | 42.38 | 60.44 | 61.61 | 59.26 | 55.82 |
| ChatGLM2-6B | 42.55 | 50.99 | 50.98 | 50.80 | 48.80 |
| XuanYuan13B | 50.07 | 64.11 | 66.32 | 59.99 | 60.05 |
| Hyacinth6B | 46.19 | 63.01 | 67.78 | 59.21 | **59.40** |

In the CMMLU benchmark, we add more Chinese models for comparison. Notably, Hyacinth6B

achieved an average score that was approximately 3 points higher than ChatGPT, which is an encouraging result. However, this could possibly be attributed to the extensive training on Chinese language data that Hyacinth6B underwent. Overall, it ranked third in total score, only behind GPT-4 and XuanYuan13B[15].

**Table4 : C-eval 5-shot Results**

| Models | STEM | SocialScience | Humanities | Others | Average |
|---|---|---|---|---|---|
| GPT4 | 67.10 | 77.60 | 64.50 | 67.80 | 68.70 |
| ChatGPT | 52.90 | 61.80 | 50.90 | 53.60 | 54.40 |
| Baichuan-13B | 47.00 | 66.80 | 57.30 | 49.80 | 53.60 |
| ChatGLM2-6B | 48.60 | 60.50 | 51.30 | 49.80 | 51.70 |
| XuanYuan13B | 56.70 | 74.20 | 62.90 | 55.90 | 61.20 |
| Hyacinth6B | 48.60 | 69.45 | 61.48 | 52.34 | **56.39** |

C-eval encompasses 52 categories, ranging from junior and senior high school subjects to a series of Q&A covering fields such as society, humanities, and engineering. The results we obtained are very similar to those from CMMLU, where our model outperformed ChatGPT by 2 points in terms of score. Overall, it was ranked third in total score, just behind XuanYuan13B and GPT-4. It is worth mentioning that Hyacinth consistently scored lower in the STEM subjects, which may be attributed to a bias resulting from a smaller volume of training datasets in these areas.

Lastly, we also employed two Traditional Chinese assessment suites to gauge the capabilities of Traditional Chinese specifically. The first is the LLM-eval[16], again proposed by the National Taiwan University Miu lab, which provides a set of 80 questions. These questions are answered by test model A and test model B, respectively, and the answers are then compared and scored by using a third-party, commercially recognized robust large language model as the judge; in this case, we choose GPT-4 as judge. This evaluation method tallies the total score to assess which model performs better. Furthermore, this evaluation is tailored to the Traditional Chinese, which serves as an additional check on the stability of Hyacinth's output in Traditional Chinese. We choose Taiwan LLM v2.0.1 chat 7B as a competitor, and the result is shown in Figure 1.

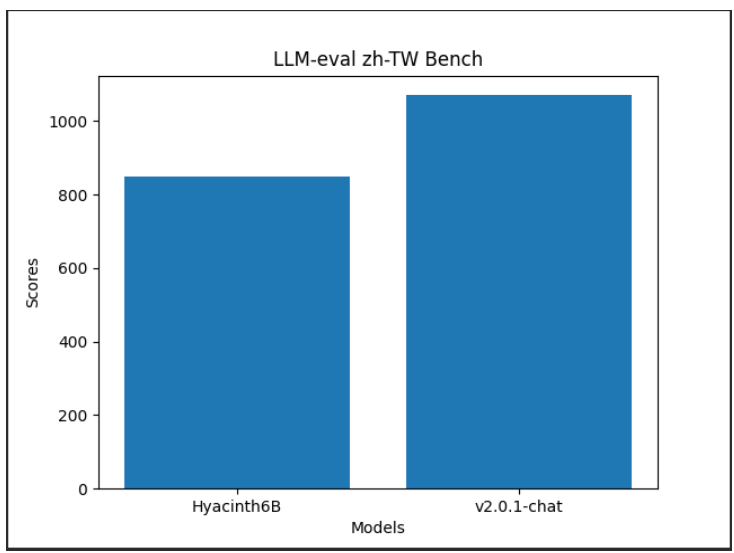

Figure1: LLM-eval result

Another evaluation suite used was TC-eval[10], provided by MediaTek Research, which similarly contains knowledge assessments across various domains, along with their curated TMMLU, the Traditional Chinese version of MMLU. The content of the questions is more inclined towards Taiwanese culture and common knowledge. For detailed scoring results, please refer to Table 5.

Table5 : TC-eval Results

| Task Name | Model7C Chat | TaiwanLLMv1.0 | ChatGPT | Hyacinth6B |
| --- | --- | --- | --- | --- |
| **XSum_TC_5k** | 0.138 | 0.004 | 0.119 | **0.152** |
| **DRCD** | 0.770 | 0.724 | **0.783** | 0.686 |
| **FGC** | **0.380** | 0.320 | 0.360 | 0.340 |
| **TTQA** | 0.650 | 0.533 | 0.747 | **0.747** |
| **IMDB_TC** | 0.915 | 0.928 | 0.940 | **0.948** |
| **PenguinsinTable_TC** | 0.268 | 0.221 | **0.402** | 0.288 |
| **TMMLU Avg.** | 0.399 | 0.304 | **0.521** | 0.272 |

The results from the TC-eval indicate that Hyacinth6B achieved the best scores in XSum_TC_5k, TTQA, and IMDB_TC. However, its performance in the TMMLU category was less impressive, suggesting that there may be room for improvement in Hyacinth6B's complex understanding, also known as the Chain of Thought (COT) aspect[17]. Overall, the performance is considered acceptable.

5. **Conclusion and future works**

In conclusion, Hyacinth6B has demonstrated that it is fully capable of handling simpler tasks, but its performance on multi-layered problem comprehension may be weaker. Assessment scores reveal that Hyacinth6B excels in social sciences while it exhibits weaker results in STEM subjects. However, this tendency towards weaker STEM performance seems to be a common issue among most open-source models, possibly due to insufficient datasets or other factors. Encouragingly, in certain evaluations, Hyacinth6B surpassed ChatGPT's scores, which not only motivates our team but also prompts us to consider whether training LLMs as domain-specific agents could be a novel research direction.

During testing phases, Hyacinth6B occasionally exhibited catastrophic forgetting[18] and misunderstandings of user queries; fortunately, such instances were not frequent. There remains substantial room for improvement in Hyacinth6B. The current results have only been achieved through supervised fine-tuning. In the future, we plan to explore the use of techniques such as reinforcement learning to enhance Hyacinth6B's performance. Emerging methodologies like Direct Preference Optimization (DPO)[19] present promising solutions, and we are considering adopting this approach to optimize Hyacinth6B further. This research journey has underscored the vast potential of language models and the continuous need for innovation in training methodologies.

6. **References**

1. Liu, Y., et al., *Summary of chatgpt-related research and perspective towards the future of large language models.* Meta-Radiology, 2023: p. 100017.

2. Ouyang, L., et al., *Training language models to follow instructions with human feedback.* Advances in Neural Information Processing Systems, 2022. **35**: p. 27730-27744.

3. Du, Z., et al. *GLM: General Language Model Pretraining with Autoregressive Blank Infilling*. 2022.

4. Zeng, A., et al., *Glm-130b: An open bilingual pre-trained model.* arXiv preprint arXiv:2210.02414, 2022.

5. Hu, E.J., et al., *Lora: Low-rank adaptation of large language models.* arXiv preprint arXiv:2106.09685, 2021.

6. Touvron, H., et al., *Llama: Open and efficient foundation language models.* arXiv preprint arXiv:2302.13971, 2023.

7. Liu, X., et al., *GPT understands, too.* AI Open, 2023.

8. Lin, Y.-T. and Y.-N. Chen, *Taiwan LLM: Bridging the Linguistic Divide with a Culturally Aligned Language Model.* arXiv preprint arXiv:2311.17487, 2023.

9. Ennen, P., et al., *Extending the Pre-Training of BLOOM for Improved Support of Traditional Chinese: Models, Methods and Results.* arXiv preprint arXiv:2303.04715, 2023.

10. Hsu, C.-J., et al., *Advancing the Evaluation of Traditional Chinese Language Models: Towards a Comprehensive Benchmark Suite.* arXiv preprint arXiv:2309.08448, 2023.

11. Hendrycks, D., et al., *Measuring Massive Multitask Language Understanding.* Proceedings of the International Conference on Learning Representations (ICLR), 2021.

12. Li, H., et al., *CMMLU: Measuring massive multitask language understanding in Chinese*. 2023.

13. Huang, Y., et al., *C-eval: A multi-level multi-discipline chinese evaluation suite for foundation models.* arXiv preprint arXiv:2305.08322, 2023.

14. OpenAI, et al., *GPT-4 Technical Report.* arXiv preprint arXiv:2303.08774, 2023.

15. Zhang, X. and Q. Yang. *Xuanyuan 2.0: A large chinese financial chat model with hundreds of billions parameters*. in *Proceedings of the 32nd ACM International Conference on Information and Knowledge Management*. 2023.

16. Lin, Y.-T. and Y.-N. Chen. *LLM-Eval: Unified Multi-Dimensional Automatic Evaluation for Open-Domain Conversations with Large Language Models*. 2023. Toronto, Canada: Association for Computational Linguistics.

17. Wei, J., et al., *Chain-of-thought prompting elicits reasoning in large language models.* Advances in Neural Information Processing Systems, 2022. **35**: p. 24824-24837.

18. Kirkpatrick, J., et al., *Overcoming catastrophic forgetting in neural networks.* Proceedings of the national academy of sciences, 2017. **114**(13): p. 3521-3526.

19. Rafailov, R., et al., *Direct preference optimization: Your language model is secretly a reward model.* arXiv preprint arXiv:2305.18290, 2023.